\documentclass[11pt]{article}

\usepackage[preprint]{acl}

\usepackage{times}
\usepackage{latexsym}
\usepackage{makecell}
\usepackage{booktabs}
\usepackage[T1]{fontenc}

\usepackage[utf8]{inputenc}
\usepackage{amsmath}
\usepackage{microtype}

\usepackage{inconsolata}

\usepackage{graphicx}
\usepackage{hyperref}
\usepackage{tcolorbox}
\usepackage{tabularx}

\title{Politics of Feelings: Emotional Expression and Legislative Effectiveness in the U.S. Congress
 }

\author{Segun Aroyehun \\
  University of Konstanz \\
  \texttt{segun.aroyehun@uni-konstanz.de}  
  }

\begin{document}
\maketitle
\begin{abstract}
Emotions are a pervasive feature of political communication, yet existing research has focused primarily on describing patterns of emotional expression rather than examining whether they are associated with consequential legislative outcomes. We address this gap by investigating the expression and correlates of discrete emotions in more than 1.7 million speeches delivered in the U.S. Congress between 1973 and 2024. Using a transformer-based emotion classifier, we measure eight discrete emotions: anger, fear, disgust, sadness, joy, enthusiasm, pride, and hope.
We examine how these emotions vary over time, across policy topics, legislator characteristics, and their relationship with legislative effectiveness.
We find that congressional speeches are becoming emotionally expressive over time.
Emotional expression also varies systematically across policy domains and ideological positioning of legislators.
Notably, the relationship between emotional expression and legislative effectiveness depends on the specific emotions expressed: enthusiasm and pride are positively associated with effectiveness, whereas anger exhibits a negative association. Emotional valence and emotional diversity are positively associated with legislative effectiveness, while emotional intensity is negatively associated with legislative effectiveness.
These findings demonstrate that computationally derived measures of discrete emotions can provide insight into affective dimensions of legislative speeches and facilitate our understanding of how legislators communicate, interact, and perform within democratic institutions.

\end{abstract}

\section{Introduction}
Political discourse is often emotional. Emotions shape attention, persuasion, and political decision-making, making them fundamental components of democratic politics rather than merely rhetorical devices \citep{marcus2000emotions}. Contemporary theories of deliberative democracy likewise argue that emotions are not antithetical to reason but can form part of democratic deliberation \citep{neblo2020impassioned}.

Recent work examines emotional expression in legislative debate across several parliamentary systems. Studies of parliamentary speeches in the United Kingdom and Germany show that emotional expression varies systematically with political, institutional, and demographic characteristics, including electoral incentives, majority status, policy issues, and speaker attributes \citep{Rheault2016ExpressionsOA, rheault2016measuring,osnabrugge2021playing,valentim2023does,yildirim2025emotions}. 
In contrast, evidence from the U.S. Congress remains more limited. Computational studies have examined broad features of congressional speech, including emotion-reason balance \citep{gennaro2022emotion} and epistemic characteristics \citep{aroyehun2025computational}. Much less is known about how specific emotions vary over time, across policy issues, between parties, and across legislators.

To investigate these questions, we analyse 1.7 million congressional speeches spanning 1973 and 2024 using a transformer-based emotion classifier that estimates the occurrence of eight discrete emotions: anger, fear, disgust, sadness, joy, enthusiasm, pride, and hope. While these discrete emotions constitute our primary focus, we also examine three broader emotional characteristics namely emotional valence, emotional intensity, and emotional diversity to evaluate whether they provide complementary explanatory power beyond individual emotions.

This study addresses four research questions that progress from describing emotional expression to evaluating its behavioural and institutional relevance.
\textbf{RQ1.} How have discrete emotions expressed in U.S. congressional speeches evolved over time, and how do these trends differ between Democratic and Republican legislators?
\textbf{RQ2.} How does the expression of discrete emotions vary across policy domains in U.S. congressional speeches?
\textbf{RQ3.} How does emotional expression by legislators vary according to individual demographic characteristics and socio-political factors?
\textbf{RQ4.} To what extent do discrete emotions and broader emotional characteristics of emotional valence, emotional intensity, and emotional diversity explain variation in the legislative effectiveness of members of Congress?

This paper makes two contributions. First, it shows that discrete emotions provide a richer account of congressional speech by revealing distinct temporal, partisan, policy-specific, and legislator-level patterns. Second, it examines whether measures of discrete emotions in text are associated with member-level legislative effectiveness, thereby linking emotional expression to legislative effectiveness of congressional members.

\section{Related Work}

\subsection{Emotions in political communication}

Prior research suggests that emotions serve distinct functions in political communication, shaping attention, persuasion, judgment, participation, and public opinion \citep{marcus2000emotions}. Functional accounts of emotion similarly argue that emotions guide different forms of appraisal, action, and social response \citep{keltner1999functional}.

Studies on parliamentary speeches show that emotional expression varies with electoral incentives, majority status, policy issues, and speaker characteristics \citep{rheault2016measuring,Rheault2016ExpressionsOA,valentim2023does,yildirim2025emotions}. For the U.S. Congress, however, computational studies have focused mainly on broader features of legislative language, including emotion-reason balance \citep{gennaro2022emotion} and epistemic characteristics of congressional speeches \citep{aroyehun2025computational}. 
Furthermore, most research on emotion in political discourse mostly considers emotional expression as an outcome.
Research in psychology, organizational behavior, and computational social science suggests that emotional expression can also be informative about consequential outcomes. Affective expressions can shape cooperation and other group dynamics \citep{barsade2002ripple, van2009emotions}. Emodiversity, the variety and relative abundance of emotions experienced or expressed, has been linked to adaptive functioning and well-being \citep{quoidbach2014emodiversity}. Related work on emotion dynamics in text shows that temporal patterns such as emotional variability, rise, and recovery can capture outcome-relevant signals beyond aggregate emotional tone \cite{Kuppens2017,vishnubhotla2024emotion}. These studies motivate asking whether emotional patterns in congressional speeches are associated with member-level legislative effectiveness \cite{volden2014legislative,volden2018legislative}.

\subsection{Measuring emotions in text}

Computational approaches to measuring emotion in text range from lexicon-based methods to supervised classifiers including transformer-based models, and large language models. Lexical resources such as the NRC emotion lexicon \cite{mohammad-turney-2010-emotions} and LIWC \cite{boyddevelopment2022} have enabled measurement of affective language across large text corpora. However, lexicon-based methods typically assign emotion scores from fixed word lists and often do not account for the context in which words are used which may result in less robust estimates \cite{jaidka2020estimating}.
Recent supervised approaches based on transformer models address this limitation by using contextual representations of language and enabling the measurement of discrete emotions at scale \cite{demszky-etal-2020-goemotions,barbieri-etal-2020-tweeteval,aroyehun2023leia}. Although large language models have expanded the toolkit for text analysis, task-specific supervised models can perform better for emotion classification in text \cite{aroyehun2023leia}. In this study, we apply a transformer-based model to identify eight discrete emotions in transcripts of U.S. congressional speeches.

\section{Methods}

\subsection{Data}

We analyze speeches from the Congressional Record covering the 93rd through the 118th U.S. Congresses (1973 to 2024). The Congressional Record provides the official transcript of floor proceedings in both the House of Representatives and the Senate and has become a standard source for studying the evolution of congressional rhetoric over time \cite{gennaro2022emotion,aroyehun2025computational, aroyehun2026epistemic, card2022computational}. 

Following the pre-processing procedure developed in previous work \cite{aroyehun2026epistemic}, we remove duplicate records, speeches with missing metadata, and speeches containing fewer than ten words. We also identify and exclude exclusively procedural interventions (e.g., requests for unanimous consent, adjournment motions, and speeches delivered by presiding officers such as the Speaker or Clerk) using a zero-shot Llama 3 model. The final corpus contains 1,766,113 speech transcripts.

Figure~\ref{fig:speech_count} (in the Appendix) presents the number of speeches in the corpus for each Congressional session, indexed by its starting year. Although the number of speeches varies across sessions, every session contains at least 20,000 speeches, providing substantial coverage for a robust and reliable analysis.

\subsection{Measures}

\subsubsection{Discrete emotions}

We quantify emotional expression using a transformer-based multilabel emotion classifier \cite{widmann2023creating} that assigns continuous scores for eight discrete emotions: anger, fear, disgust, sadness, joy, enthusiasm, pride, and hope. The multilabel architecture allows multiple emotions to co-occur within the same speech rather than forcing each text into a single emotion category. 

The selected emotion categories provide a balanced representation of emotions with negative (anger, fear, disgust, and sadness) and positive (joy, enthusiasm, pride, and hope) valence. More importantly, they distinguish emotional states that differ in their underlying appraisals and behavioural implications. For example, anger and fear both convey negative affect but differ in their associations with threat, certainty, blame, and political action. Likewise, joy, enthusiasm, pride, and hope capture distinct forms of positive emotional expression. Measuring these emotions separately therefore preserves information that aggregate sentiment measures might obscure.

The classifier generates one score for each emotion for every speech. We standardize each emotion score across the full corpus before further analyses using z-transformation. The standardization preserves the underlying distribution of each emotion while placing all emotion categories on a common scale, thereby facilitating direct comparisons across emotions. 

We evaluate the face validity of the emotion measures in three ways. First, we inspect speeches receiving the highest scores for each emotion to determine whether they express substantively plausible examples of the corresponding emotional category. Second, we examine the correlation structure among the eight emotions to assess whether theoretically related emotions display plausible patterns of association. Third, we manually annotate a sample of high- and low-scoring speeches for each emotion to evaluate whether the classifier's predictions correspond to human judgments. Appendix \ref{sec:classifier_eval} presents these validation analyses, which provide quantitative and qualitative support for the validity of the emotion measures.

\subsubsection{Broader emotional characteristics}

Although discrete emotions constitute the primary focus of this study, researchers often summarize emotional expression using broader affective characteristics \cite{Kuppens2017, teodorescu2023language}. We therefore derive three complementary measures: emotional valence, emotional intensity, and emotional diversity.
We operationalize emotional valence as the difference between the aggregate scores of emotions with positive and negative valence. This measure captures the overall balance between positive and negative emotional expression.
We measure emotional intensity as the sum of the eight emotion scores. This measure reflects the overall magnitude of emotional expression regardless of whether it is positive or negative.

Finally, we quantify emotional diversity as the entropy of the normalized emotion score distribution. Higher entropy indicates that emotional expression is distributed across multiple emotions, whereas lower entropy indicates that a small number of emotions are more prominent in the distribution.
Together, these measures complement the discrete emotion scores. Whereas the discrete measures identify which specific emotions legislators express, valence, intensity, and diversity summarize broader characteristics of their communication.

\subsubsection{Legislative effectiveness}

We measure legislative effectiveness using the Legislative Effectiveness Score (LES) developed in prior work \cite{volden2014legislative,volden2018legislative}. The LES is a member-Congress-level measure that captures how successful each legislator is at advancing sponsored bills through the legislative process, from introduction to final passage. The dataset covers congressional sessions from 1973 to 2024 and allows us to link emotional expression in congressional speeches to variation in legislative effectiveness across members and congressional sessions.

\subsubsection{Policy topics}

We examine policy-specific variation using the Comparative Agendas Project (CAP) policy taxonomy. The CAP framework provides a standardized taxonomy of major policy domains, allowing researchers to compare issue attention consistently across time and institutions.
We assign each speech to one of 21 major CAP policy topics using a transformer-based topic classifier developed in previous work \cite{aroyehun2025computational}. The classifier achieves a macro-averaged F1 score of 0.857 on held-out test data. We use the resulting topic assignments for two purposes. First, we examine how emotional expression varies across substantive policy domains. Second, we include policy topics as control variables in the regression models because emotional expression likely differs across issue areas.

\subsection{Analytic strategy}

Our analytical strategy addresses four research questions concerning the temporal evolution, policy-specific variation, correlates, and institutional consequences of emotional expression in Congressional speeches.
To address the first research question on temporal evolution, we aggregate standardized emotion scores by congressional session and political party and calculate the mean value for each emotion. We visualize the patterns using the Congress starting year as the temporal index. We then assess monotonic temporal trends using the Mann--Kendall trend test \cite{mann1945nonparametric, Hussain2019pyMannKendall} and estimate the magnitude of change using Sen's slope \cite{sen1968estimates}. Since we conduct multiple hypothesis tests across emotion categories, we adjust statistical significance using the Benjamini--Hochberg false discovery rate procedure \cite{benjamini1995controlling}.

To address the second research question on potential variation across policy domains, we aggregate emotion scores by CAP policy topic and calculate mean standardized emotion scores for each policy domain. We summarize these patterns using a topic-by-emotion heatmap and test whether emotion distributions differ across policy areas using the Kruskal--Wallis test. We further examine topic-specific temporal patterns as a supplementary analysis to assess whether policy-specific emotional patterns remain stable over time.

To address the third research question on correlates of emotions, we estimate separate linear mixed-effects models for each emotion using speech-level emotion scores as the dependent variable. We include random intercepts for legislators to account for repeated observations within individuals. The models include demographic, political, and institutional characteristics as explanatory variables, while controlling for policy topic and including Congress fixed effects to account for secular temporal changes shared across legislators.

Finally, we investigate whether emotional expression predicts legislative effectiveness. We estimate mixed-effects regression models with legislative effectiveness as the dependent variable. We first estimate separate models for each discrete emotion and subsequently estimate models using emotional valence, emotional intensity, and emotional diversity as alternative summary measures. Comparing these models allows us to assess whether broader emotional characteristics can also explain variation in legislative effectiveness.

\section{Results}

\subsection{RQ1: Temporal trends in expression of emotions }
\begin{figure*}[htbp]
    \centering
    \includegraphics[width=\textwidth]{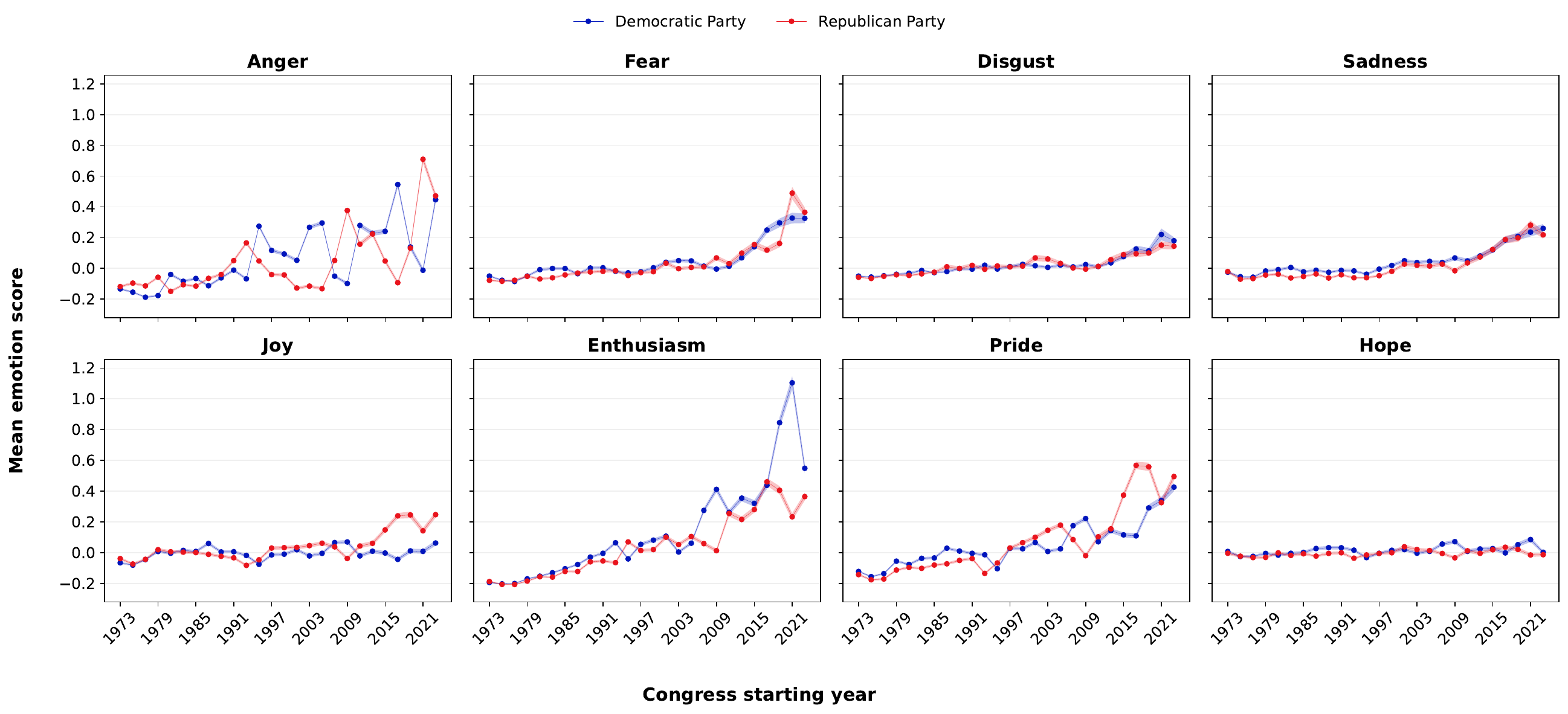}
    \caption{Temporal trends in mean standardized emotion scores by party over time (Congressional sessions). Each dot represents Congress-level mean emotion scores, while shaded ribbons indicate 95\% confidence intervals computed from the standard error of the mean within each Congress–party group. The large number of speeches contributing to each estimate results in very narrow confidence intervals, which may be indistinguishable from the trend lines.}
    \label{fig:emotion_trends_party}
\end{figure*}

Figure~\ref{fig:emotion_trends_party} shows Congress-level mean emotion scores for Democratic and Republican legislators starting from the 93rd (1973) to the end of 118th (2024) Congress. Across the eight discrete emotions, several consistent temporal patterns emerge.
First, both parties exhibit increase in emotional expression over time. 
The Mann–Kendall tests confirm statistically significant increasing monotonic trends for nearly all emotions after controlling for multiple testing (see Table~\ref{tab:mk_test_stats} in the Appendix), with the only notable exception being joy among Democratic legislators.
Notable rise occurs for enthusiasm and pride, both of which show pronounced upward trajectories for Democratic and Republican legislators. Democratic legislators display a particularly sharp increase in enthusiasm during the final Congressional session of the study period, whereas Republican legislators exhibit a steadier increase. Pride likewise increases substantially for both parties, although Republican legislators express higher levels during much of the last decade.
Among the emotions with negative valence, fear, disgust, and sadness also increase steadily throughout the period. Fear displays one of the clearest monotonic trends, with both parties following remarkably similar trajectories. Disgust and sadness likewise rise gradually over time.
Anger exhibits temporal volatility. Although both parties show a significant increase, session-level averages fluctuate considerably, particularly during the last two decades. Democratic legislators display pronounced peaks during the early 2000s and again around 2017, while Republican legislators exhibit higher anger scores during the most recent congressional sessions (2021–2024). 
This greater variability suggests that expression of anger by legislators may be more contingent on the prevailing political context. 

Positive emotions exhibit more heterogeneous patterns. While enthusiasm and pride increase consistently, joy remains comparatively stable among Democrats, consistent with the non-significant Mann–Kendall trend, but increases gradually among Republicans. In contrast, hope remains comparatively stable for both parties despite exhibiting a significant monotonic trend.
Overall, the trajectories indicate that the expression of most discrete emotions in congressional speeches has increased over the past five decades. 
Emotions with both positive and negative valence increase over time, suggesting a broad intensification of emotional expression rather than a simple shift towards more negative political rhetoric. Moreover, the largely parallel trajectories observed for Democrats and Republicans suggest that these trends reflect broader changes in congressional communication. 

\subsection{RQ2: Emotional expression across policy topics}
\begin{figure*}[htbp]
    \centering
    \includegraphics[width=\textwidth]{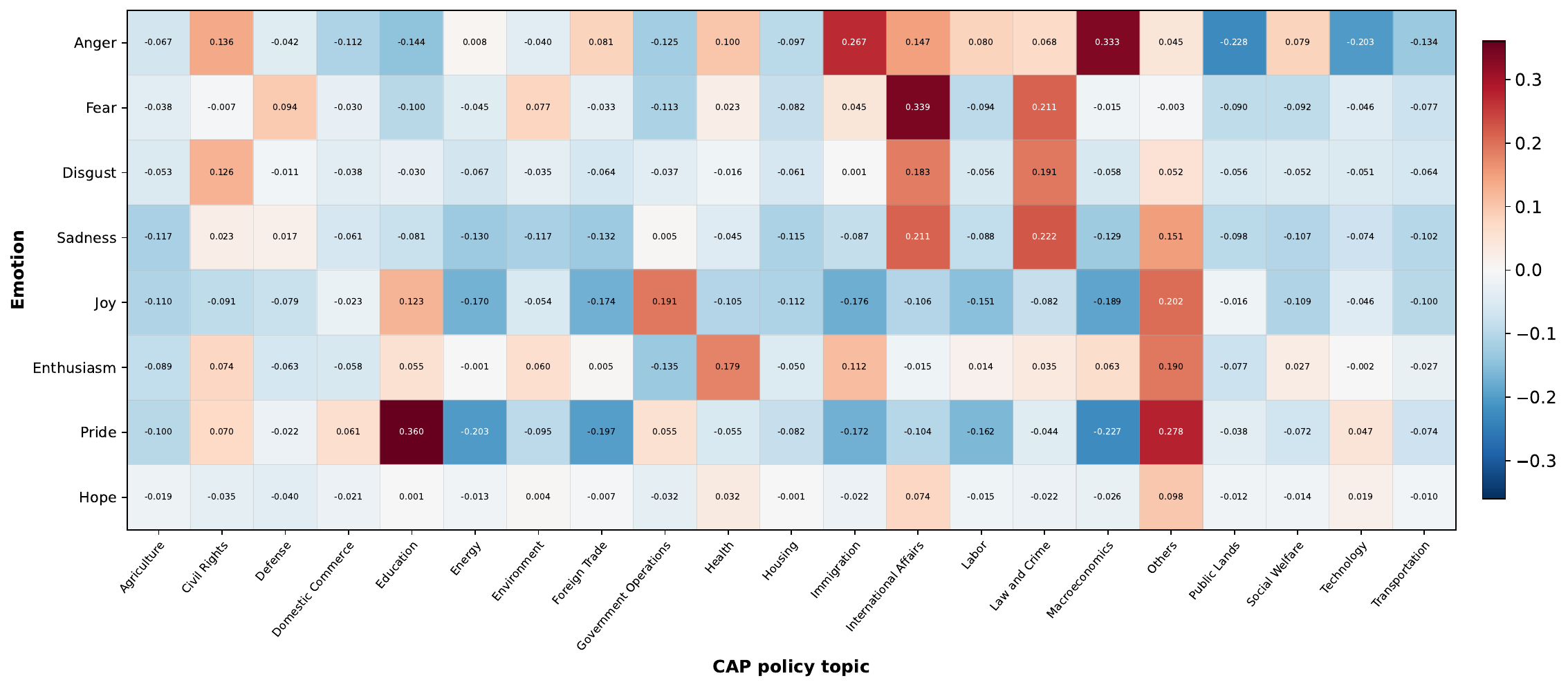}
    \caption{Emotional profiles across policy domains. Heatmap of mean standardized emotion scores across Comparative Agendas Project (CAP) policy domains. Each cell represents the average standardized score for an emotion within a policy domain. Positive values indicate relatively higher expression of the corresponding emotion, whereas negative values indicate relatively lower expression.}
    \label{fig:emotions_topics_heatmap}
\end{figure*}

Figure~\ref{fig:emotions_topics_heatmap} depicts the aggregate emotion scores across the 21 major CAP policy topics. Emotional expression varies substantially across policy domains, indicating that the emotional content of congressional speeches differs systematically according to the policy area under discussion. Figure~\ref{fig:emotions_trend_bytopic}  (in the Appendix) shows that these topic-specific patterns remain mostly stable over time. The Kruskal--Wallis tests further indicate that the distributions of all eight emotion scores differ significantly across policy topics (Table~\ref{tab:kw_test_emotion_topics} in the Appendix).
Several policy domains exhibit relatively high scores across multiple emotions with negative valence. Speeches focusing on International Affairs show the highest average fear scores and also elevated anger, disgust, and sadness. Law and Crime likewise exhibits relatively high fear, disgust, and sadness, while speeches on Immigration record the highest mean anger score among the policy topics.
Positive emotions display equally distinct policy-specific patterns. Speeches on Education exhibit the highest average pride score together with elevated joy and enthusiasm. The Health domain is characterized by relatively high enthusiasm, whereas Government Operations exhibits comparatively high joy. The residual Others category also records relatively high scores across several positive emotions.
Other policy domains exhibit comparatively lower emotional expression. Speeches on Agriculture, Housing, Transportation, Technology, and Public Lands generally display lower scores across most emotion categories, while Macroeconomics combines relatively high anger with comparatively low joy and pride.
Overall, the results demonstrate that emotional expression varies systematically across policy domains.

\subsection{RQ3: Correlates of emotions}

\begin{figure}[htbp]
    \centering
    \includegraphics[width=0.47\textwidth]{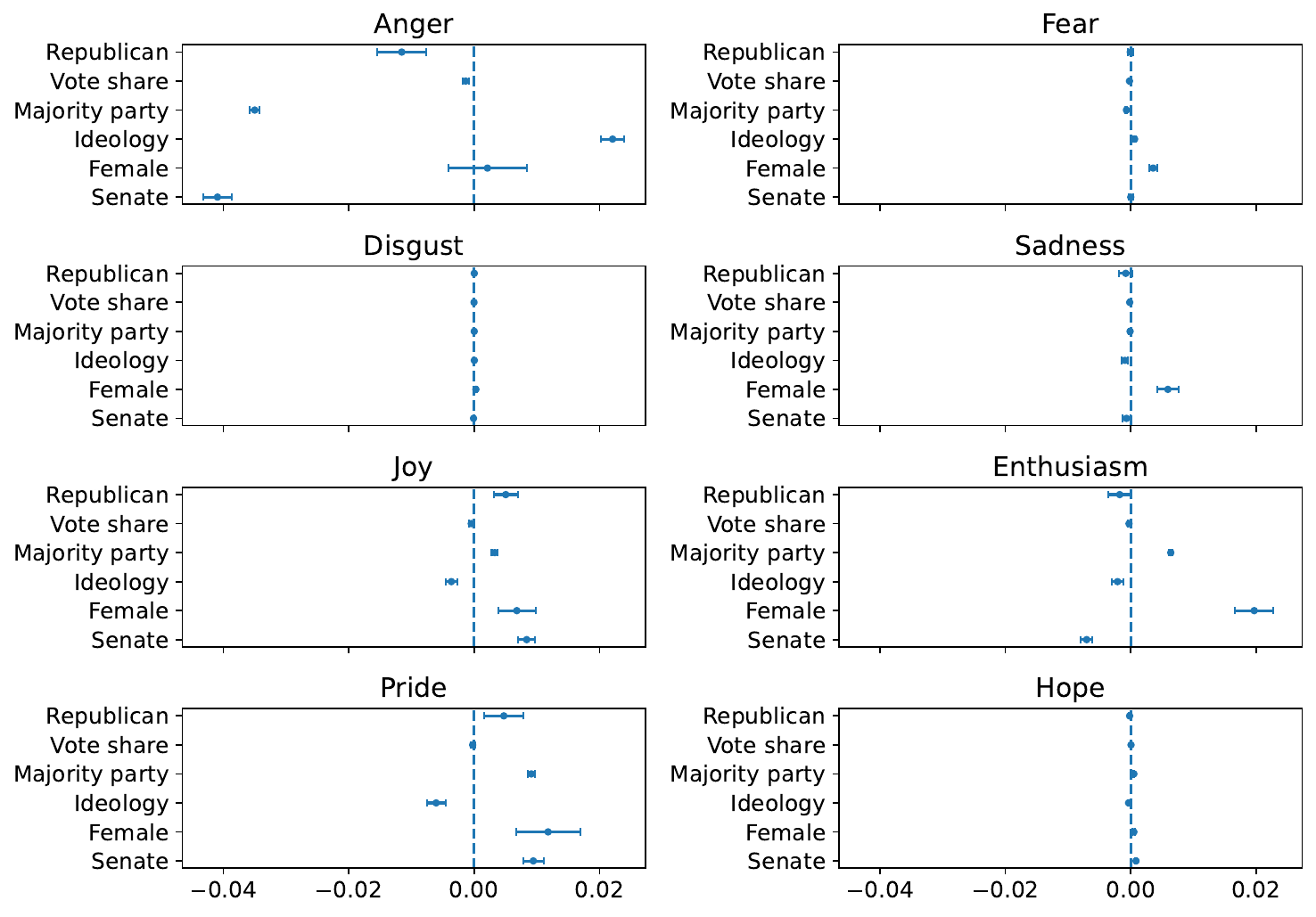}
    \caption{Correlates of emotional expression in congressional speeches. Each panel corresponds to a separate mixed-effects model for one emotion. Points denote coefficient estimates and horizontal bars denote 95\% confidence intervals. All models use identical specifications and control variables.}
    \label{fig:emotion_correlates}
\end{figure}

Figure~\ref{fig:emotion_correlates} summarizes the estimated associations between demographic and political characteristics of members of Congress and the expression of the eight discrete emotions. We report he full coefficient estimates in Appendix Tables~\ref{tab:positive_emotion_correlates} and ~\ref{tab:negative_emotion_correlates}.

\paragraph{Joy} Female legislators (b=0.007, 95\% CI [0.004, 0.010], p < 0.001), Senators (b=0.008, 95\% CI [0.007, 0.010], p < 0.001), members of the majority party (b=0.003, 95\% CI [0.003, 0.004], p < 0.001), and legislators of the Republican party (b=0.005, 95\% CI [0.003, 0.007], p < 0.001) express higher levels of joy. Ideology score predicts lower expression of joy (b=-0.004, 95\% CI [-0.005, -0.003], p < 0.001).

\paragraph{Enthusiasm}Female legislators express higher levels of enthusiasm (b=0.020, 95\% CI [0.017, 0.023], p<0.001). Members of the majority party also express higher levels of enthusiasm (b=0.006, 95\% CI [0.006, 0.007], p<0.001), whereas Senators express lower levels (b=-0.007, 95\% CI [-0.008, -0.006], p<0.001). Ideology score predicts lower levels of enthusiasm (b=-0.002, 95\% CI [-0.003, -0.001], p<0.001).

\paragraph{Pride} Female legislators (b=0.012, 95\% CI [0.007, 0.017], p<0.001), Senators (b=0.009, 95\% CI [0.008, 0.011], p<0.001), members of the majority party  (b=0.009, 95\% CI [0.009, 0.010], p<0.001), and legislators of the Republican party (b=0.005, 95\% CI [0.002, 0.008], p<0.001) express higher levels of pride. Ideology score predicts lower levels of pride (b=-0.006, 95\% CI [-0.008, -0.005], p<0.001).

\paragraph{Hope} Senators express higher levels of hope (b=0.001, 95\% CI [0.001, 0.001], p<0.001). Although female and Republican legislators, ideology score, and members of the majority party also exhibit statistically significant associations, their estimated coefficients are close to zero.

\paragraph{Anger} 

Ideology score predicts greater expression of anger (b=0.022, 95\% CI [0.020, 0.024], p<0.001). Senators (b=-0.041, 95\% CI [-0.043, -0.039], p<0.001), members  of the majority party (b=-0.035, 95\% CI [-0.036, -0.034], p<0.001), and legislators of the Republican party  (b=-0.012, 95\% CI [-0.015, -0.008], p<0.001) express lower levels of anger. Legislators elected with a higher vote share also express less anger (b=-0.001, 95\% CI [-0.002, -0.001], p<0.001).

\paragraph{Fear} Female legislators express higher levels of fear (b=0.004, 95\% CI [0.003, 0.004], p<0.001). Ideology score also predicts greater expression of fear (b=0.001, 95\% CI [0.000, 0.001], p<0.001), whereas members of the majority party express lower levels of fear (b=-0.001, 95\% CI [-0.001, -0.001], p<0.001).

\paragraph{Disgust} All estimated coefficients are close to zero. 
\paragraph{Sadness}Female legislators express higher levels of sadness (b=0.006, 95\% CI [0.004, 0.008], p<0.001). Ideology score predicts lower expression of sadness (b=-0.001, 95\% CI [-0.001, -0.000], p<0.001).

\subsection{RQ4: Emotions as predictor of legislative effectiveness }

\begin{figure}
    \centering
    \includegraphics[width=0.45\textwidth]{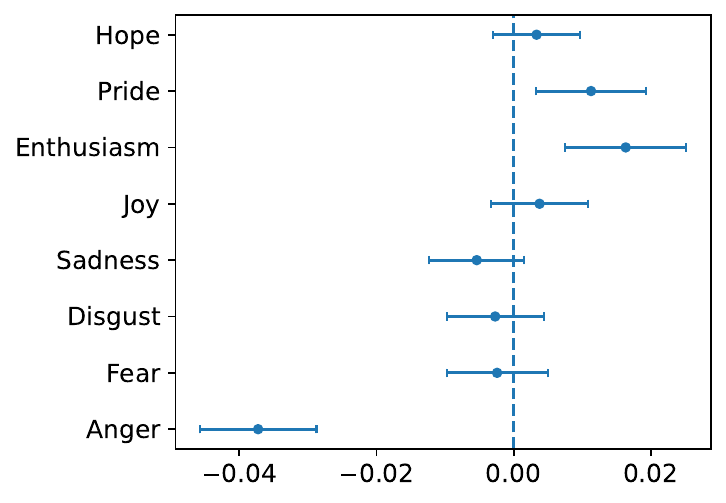}
    \caption{Emotions as predictors of legislative effectiveness. Points denote coefficient estimates and horizontal bars denote 95\% confidence intervals from separate mixed-effects models, with one model estimated for each emotion. All models use identical specifications and control variables.}
    \label{fig:emotions_les}
\end{figure}

Figure~\ref{fig:emotions_les} provides a summary of the estimated associations between emotional expression of legislators and their legislative effectiveness score. 
The analyses reveal heterogeneity across the eight discrete emotions. Enthusiasm (b=0.016, 95\% CI [0.007, 0.025], p<0.001) and pride (b=0.011, 95\% CI [0.003, 0.019], p < 0.01) have a positive association with legislative effectiveness. In contrast, anger is the only emotion that has a statistically significant negative association with legislative effectiveness (b=-0.037, 95\% CI [-0.046, -0.029], p<0.001).

To examine whether broader affective characteristics provide additional insight, we next estimated analogous models using i.  emotional valence and emotional diversity, and ii. emotional intensity and emotional diversity in place of the individual emotions (Table~\ref{tab:les_broader_emotions} in the Appendix). Comparing these models with the discrete emotion analyses allows us to evaluate whether aggregated emotional characteristics also capture meaningful variation. 
The results show that emotional diversity is positively associated with legislative effectiveness in both model specifications (b=0.026, 95\% CI [0.017, 0.034], p<0.001; b=0.031, 95\% CI [0.024, 0.039], p<0.001). Emotional valence is also positively associated with legislative effectiveness (b=0.020, 95\% CI [0.011, 0.028], p<0.001), whereas emotional intensity (emotionality) is negatively associated with legislative effectiveness (b=-0.014, 95\% CI [-0.025, -0.004], p<0.01).

\section{Discussion and conclusion}
The findings suggest that discrete emotions are more than descriptive properties of legislative speeches. Across more than five decades of Congressional speeches, emotional expressions show systematic variation across time, policy domains, legislators, and legislative effectiveness. 

Together, these patterns indicate that computationally derived measures of discrete emotions can recover meaningful dimensions of legislative communication. 
Rather than simply capturing how emotional congressional speech is, they reveal how legislators evaluate political issues, position themselves relative to others, and communicate within the institutional context of Congress.
The analyses reveal that emotional expression exhibits systematic temporal and substantive structure. 
Anger, fear, disgust, and sadness increase over time, so do enthusiasm and pride, suggesting that members of Congress have become more emotionally expressive in their speeches. Equally noteworthy, Democrats and Republicans show similar temporal trajectories across most emotions. This observation aligns with recent longitudinal work \cite{aroyehun2025computational} showing similar co-movement in the epistemic characteristics of speeches. 

Emotional expression also varies systematically across policy domains. Different topics exhibit distinct emotional profiles that remain broadly stable over time, suggesting that emotional expression reflects the recurring communicative demands of legislative debate. This interpretation is reinforced by qualitative inspection of speeches with high scores. High-scoring expressions of pride involve legislators recognising colleagues, honouring constituents, celebrating local communities, or acknowledging legislative achievements rather than expressing self-focused pride. For example, ``I am grateful for the contributions of our staff...the many Members who were involved in shaping this bill. It is another accomplishment that we can all be proud of''.
Similarly, speeches expressing enthusiasm employ motivational language encouraging legislative action, emphasising shared goals, or projecting optimism about proposed legislation. For example, ``We can right that wrong. This is America. We can make it right ...''.
In contrast, speeches with high anger scores typically focus on criticism, attribution of responsibility, or perceived policy failures. For example, ``... the House created another monumental disaster. This socalled reform bill has all the
makings of creating further budgetary chaos in the welfare system of this Nation...''. These examples suggest that the expressed emotions function as communicative resources through which legislators perform their tasks or aim for specific objectives.

Legislative speech is interactional. Legislators not only communicate policy positions but also evaluate issues and position themselves in relation to other legislators. Research on stance and appraisal conceptualises these processes as interconnected, whereby evaluation simultaneously positions speakers towards an issue and aligns them with or distances them from others \cite{dubois2007stance, martin2005language}. The expression of discrete emotions may therefore shape both the evaluative and relational dimensions of legislative communication.

The associations with legislative effectiveness suggest that these differences are consequential. Enthusiasm and pride are positively associated with legislative effectiveness, whereas anger is negatively associated with legislative effectiveness. In an institution where legislative work depends on persuasion, coordination, and working relationships, how legislators express emotions matters.
Broader affective measures reinforce this interpretation. Emotional diversity and valence are positively associated with legislative effectiveness, whereas emotional intensity is negatively associated with legislative effectiveness. Emotional diversity (the variety and relative abundance of different emotions) has been associated with indicators of psychological and physical well-being \cite{quoidbach2014emodiversity, Grossmann2019Wise}. In legislative speeches, expressing a broader range of emotions may similarly reflect greater flexibility in communicating across different legislative contexts. 
Aggregate measures of emotion, while instructive, can obscure differences among discrete emotions with distinct communicative functions. Examining discrete emotions therefore provides a richer account of how legislators use emotional language to evaluate political issues, manage relationships, coordinate action, and perform the interactional work of democratic representation.

\section{Limitations}
Given the scope of this study, several limitations should be acknowledged.
\paragraph{Generlization to other contexts} The current study focuses on the U.S. Congress where there is availability of extensive data to enable the analyses reported in this paper.
Nevertheless, it remains an open question whether the observed patterns and relationships between emotional expression and legislative effectiveness generalize to other legislative contexts. Future work should examine the extent to which these findings apply in subnational legislatures (e.g., U.S. states) or supranational institutions (e.g., the European Parliament), as well as in legislatures operating under different institutional and party systems.

\paragraph{Causes and targets of emotions}
Although we account for the policy domain of speeches using the CAP policy topics, future work should identify causes and targets of emotional expressions. Doing so, will enable a richer understanding of the strategic and communicative functions of emotions in legislative discourse.

\paragraph{A broader range of discrete emotions} The set of discrete emotions examined in this study is necessarily selective rather than exhaustive. While we focus on emotions that are relevant to political communication and provide a balance between positive and negative emotions, an avenue for future research is to examine additional emotions such as curiosity, guilt, shame, and nostalgia.

%
%
\bibliography{custom}

\appendix{}
\section{Dataset}
\begin{figure*}[htbp]
    \centering
    \includegraphics[width=\textwidth]{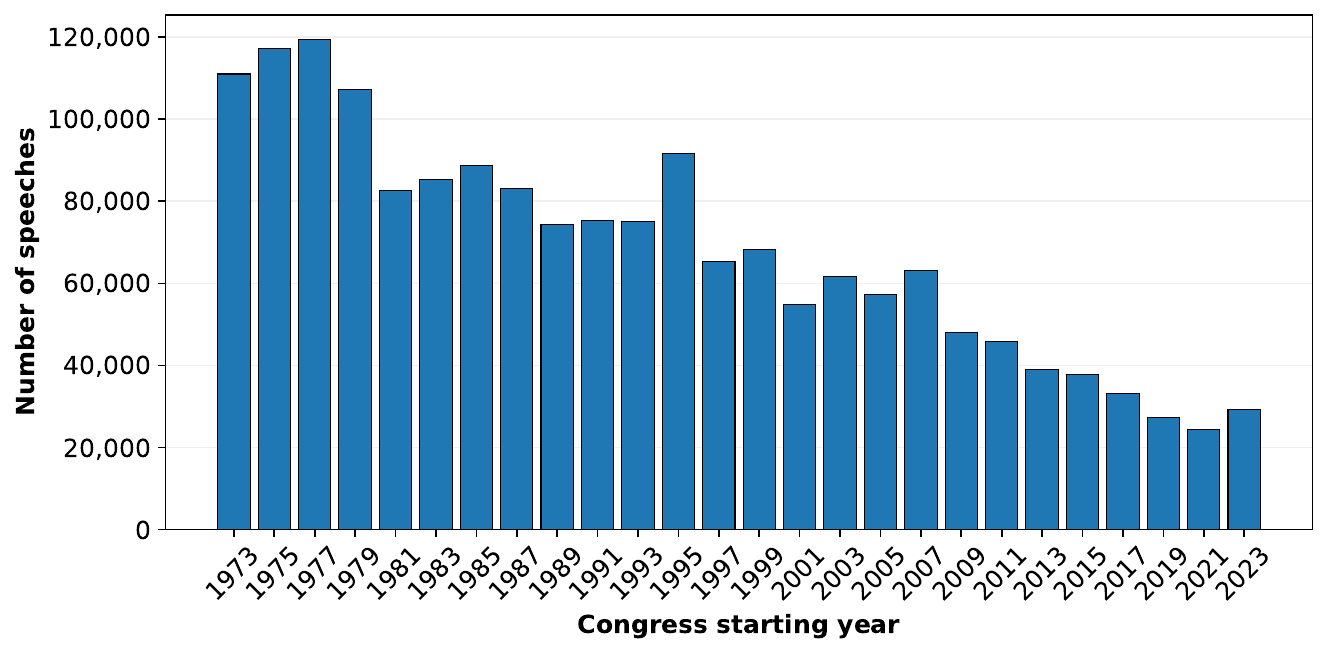}
    \caption{Number of speeches over time (Congressional sessions)}
    \label{fig:speech_count}
\end{figure*}

Figure \ref{fig:speech_count} shows the number of speeches for each Congressional session in the dataset.
Each Congressional session has at least 20,000 speeches. While corpus size varies across sessions, each period contains a sufficiently large number of observations to support reliable analyses of emotional expressions.

\section{Classifier evaluation}
\label{sec:classifier_eval}
For each emotion, we select 25 responses with the highest predicted probabilities and the 25 responses with the lowest predicted probabilities (for a total of 50). These responses are manually annotated and treated as the reference labels. We then evaluate the correspondence between classifier scores and human annotations using the area under the receiver operating characteristic curve (AUC), which assesses the extent to which higher classifier probabilities are assigned to human-labeled positive cases relative to negative cases. We use AUC because it evaluates ranking quality independently of a fixed classification threshold and is therefore well suited for continuous probabilistic scores. Similar validation approach has been used in prior work examining automatically derived continuous measures \cite{gennaro2022emotion, aroyehun2025computational}. Table \ref{tab:emotion_classifer_auc} shows the results.

\begin{figure}
    \centering
    \includegraphics[width=0.45\textwidth]{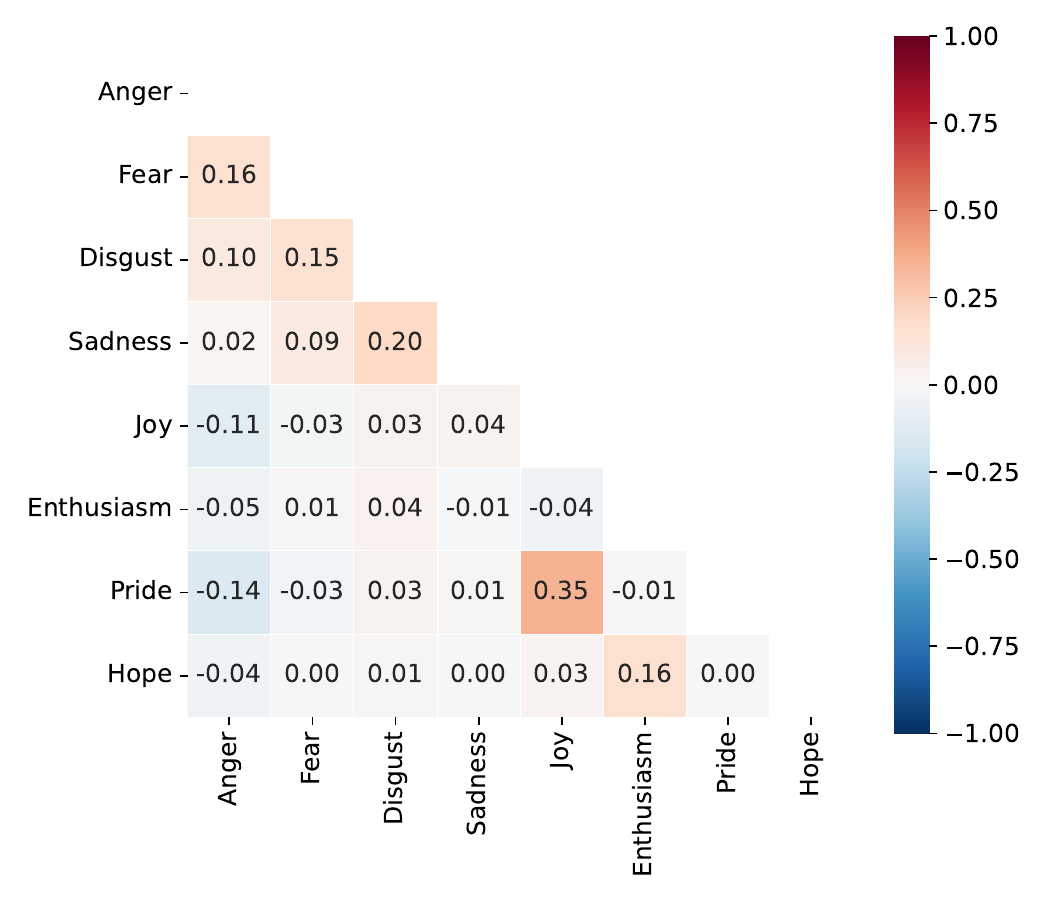}
    \caption{Co-occurrence of emotions in the predictions of the classifier model on the Congressional seeches}
    \label{fig:emotion_cooccurrence}
\end{figure}

Beyond the sample, we also assess the co-occurrence of emotions as another validity check. The correlation structure (see Figure \ref{fig:emotion_cooccurrence}) provides additional face validity. Negative emotions display modest positive associations with one another, whereas positive emotions are more differentiated. Cross-valence correlations are generally weak, suggesting that the measures capture related and distinct emotions.

\begin{table}[htbp]
    \centering
    \begin{tabular}{lrr}
\hline
 Emotion & AUC \\
\hline
Anger & 1.000  \\
Fear & 0.952  \\
Disgust & 0.922  \\
Sadness & 0.989  \\
Joy & 1.000  \\
Enthusiasm & 0.982  \\
Pride & 0.981  \\
Hope & 1.000  \\
\hline
\end{tabular}
    \caption{In-domain classifier validation}
    \label{tab:emotion_classifer_auc}
\end{table}

\clearpage

\begin{table*}[htbp]
    \centering
    \small
    \begin{tabularx}{\textwidth}{>{\bfseries}p{2.2cm}X}
        \hline
         Emotion & Example high scoring text \\
         \hline
         Anger & Mr. Speaker. today the House created another monumental disaster. This socalled reform bill has all the makings of creating further budgetary chaos in the welfare system of this Nation. The provisions in the bill are poorly thought out and represent an amalgam of bad ideas whose time has. at best. come and gone. To have such a bad bill come to the floor under a closed rule is tragedy. Extensive revision of the provisions of this bill would be needed to even begin to make some sense out of it. It is a Trojan horse. the further implementation of the liberal goal of a guaranteed annual income. It is becoming a national embarrassment to watch this Chamber destroy the last vestiges of democracy in the Congress. Time after time since the 96th Congress was called into session the ability of Members to debate bills and offer amendments has been shackled \\ \hline
         Fear & Mr. President. as the attention of our Nation and the world remains riveted on the events in Beirut. it is sobering indeed to consider the possibility of the ultimate terrorism that which puts at risk the lives not merely of dozens or hundreds of hostages but of whole cities and nations. I am speaking of the threat of nuclear terrorism a danger that has grown all the more real in recent. years. It is important to consider this danger now because it increases as terrorists win the support of radical states and the significant resources and capabilities that come with such support. This is the ultimate nightmare, one that could trigger a nuclear exchange between the superpowers, and we may drift into it unless meaningful steps are taken promptly to prevent it from happening. \\ \hline
         Sadness & They never will see them again, taken so cruelly and so quickly. Today, flags around the Capitol will remain at half-staff in honor of the victims, and we all grieve with their families. We also grieve with the community of Boulder and the people of Colorado. And we grieve with the people of Georgia and all people across the United States whose lives have been forever marred by the plague of gun violence. COVID-19 is not the only epidemic claiming innocent lives in America. Last year alone, 20,000 Americans were killed by gun violence, the highest number in almost two decades. Most of these incidents never reached the headlines, but we cannot allow ourselves to become numb to their devastation. After one of the most difficult years in American history, we all want our lives and our country to return to normal. \\ \hline
         
         Disgust & The actors will shoot and stab and beat one another to a bloody pulp. They will firebomb cars and buildings. They will enact at least 32 such gory scenes each hour--about two a minute. And they will come back and repeat these enactments tomorrow, and the next day, and the day after that. They will spend more time with your children than you do yourself. They will teach them to solve problems and to settle disputes by killing. They will convey to them that the adult world approves of this kind of behavior that it is glamorous and attractive and the way the coolest adults themselves behave. Is there a parent in this country that would accept that offer? More likely. they would call the police. They would report the man in the expensive suit for child molesting. Yet precisely this transaction happens every day in America. \\ \hline
        
         Joy & I thank the Senator very. very much. I am very grateful for that and I appreciate his cooperation prior to this. It has been a great help. \\ \hline
         Hope & I am hopeful that New Mexico. as well as many other communities across the nation. will see great benefits as a result of this legislation. I hope that we are successful at reviving the ghost towns that currently exist in many downtown areas and that they will once again come alive with prosperity.\\ \hline
         Pride & I am the first to say that nothing consequential or substantial gets done around this place without the important. hard work of the very talented and skilled staff. I am blessed on the Commerce Committee to be surrounded with people who care passionately about these issues. who work very diligently to get the best possible outcomes and results. I am grateful for the contributions of our staff and those of Senator NELSONs staff and of the many Members who were involved in shaping this bill. It is another accomplishment that we can all be proud of. With that. I yield the floor. \\ \hline
         
         Enthusiasm & We can right that wrong. This is America. We can make it right. The bill the Senate passed makes it right. This is a moral matter. It is a moral commitment. I call on the House to act without delay. I call on them to do the right thing without hesitation. I urge the Speaker: Do what is right for the American people. Do what is right for the working people of this Nation. If you do, the Nation will commend you and stand with you. Courage in the service of justice is what we need now, and I urge them toward it. I yield the floor.\\ \hline
    \end{tabularx}
    \caption{Illustrative texts with high classifier scores for each emotion category}
    \label{tab:example_high_emotion_text}

\end{table*}

\section{Illustrative examples of text with high scores based on the emotion classifier}
Table \ref{tab:example_high_emotion_text} presents examples of text with high classifier scores for each of the discrete emotions. The examples suggest that the classifiers are capturing the intended constructs (discrete emotions).

\section{Regression models}

We estimate two sets of mixed-effects regression models. First, to examine correlates of emotional expression, we model each standardized emotion score as a function of legislator demographic and political characteristics with random intercepts for legislators. Second, to examine the relationship between legislative effectiveness and discrete emotions, we model member-level LES as a function of emotion measures and the set of controls introduced in prior work on legislative effectiveness \cite{volden2014legislative,volden2018legislative}.

We specify the two models as follows:
\begin{equation*}
Emotion_{ic} = \alpha + X_{ic}\beta + T_{ic}\theta + \delta_c + u_i + \varepsilon_{ic},
\end{equation*}

\begin{equation*}
LES_{ic} = \alpha + \beta Emotion_{ic} + X_{ic}\gamma + T_{ic}\theta + \delta_c + u_i + \varepsilon_{ic},
\end{equation*}

where $i$ indexes legislators and $c$ indexes congressional sessions. $Emotion_{ic}$ denotes the standardized average emotion score for legislator $i$ in Congress $c$, and $LES_{ic}$ denotes the legislative effectiveness score measured at the member-Congress level. $\mathbf{X}_{ic}$ includes legislator-level demographic and political characteristics, $T_{ic}$ denotes CAP policy topic controls, $\delta_c$ denotes Congress fixed effects, and $u_i$ is a legislator-level random effect.

Tables \ref{tab:positive_emotion_correlates} and \ref{tab:negative_emotion_correlates} show the output of the regression model assessing the relationship between emotions and covariates.
Tables \ref{tab:les_negative_emotions}, \ref{tab:les_negative_emotions}, and \ref{tab:les_broader_emotions} show the output of the regression model assessing the relationship between legislative effectiveness and discrete emotions as well as aggregate emotional characteristics.

\section{Statistical tests for trends in discrete emotions}
Table \ref{tab:mk_test_stats} for the mk trend test. 
Table \ref{tab:mk_test_stats} reports Mann--Kendall trend tests, Sen's slope estimates, and FDR-adjusted significance levels for monotonic temporal trends in discrete emotion expression.

\begin{table*}[htbp]
    \centering
    \begin{tabular}{llrlrrrr}
\hline
 Party & Emotion & N & Trend & $\tau$ & Sen's slope & p & p (fdr) \\
\hline
Democratic Party & Enthusiasm & 26 & increasing & 0.877 & 0.030 & 0.000 & 0.000 \\
Democratic Party & Disgust & 26 & increasing & 0.834 & 0.006 & 0.000 & 0.000 \\
Democratic Party & Sadness & 26 & increasing & 0.772 & 0.009 & 0.000 & 0.000 \\
Democratic Party & Pride & 26 & increasing & 0.754 & 0.016 & 0.000 & 0.000 \\
Democratic Party & Fear & 26 & increasing & 0.735 & 0.009 & 0.000 & 0.000 \\
Democratic Party & Anger & 26 & increasing & 0.557 & 0.020 & 0.000 & 0.000 \\
Democratic Party & Hope & 26 & increasing & 0.391 & 0.002 & 0.005 & 0.006 \\
Democratic Party & Joy & 26 & no trend & 0.243 & 0.002 & 0.086 & 0.086 \\
Republican Party & Fear & 26 & increasing & 0.877 & 0.010 & 0.000 & 0.000 \\
Republican Party & Enthusiasm & 26 & increasing & 0.822 & 0.023 & 0.000 & 0.000 \\
Republican Party & Pride & 26 & increasing & 0.815 & 0.023 & 0.000 & 0.000 \\
Republican Party & Disgust & 26 & increasing & 0.742 & 0.007 & 0.000 & 0.000 \\
Republican Party & Sadness & 26 & increasing & 0.717 & 0.009 & 0.000 & 0.000 \\
Republican Party & Joy & 26 & increasing & 0.612 & 0.008 & 0.000 & 0.000 \\
Republican Party & Anger & 26 & increasing & 0.434 & 0.013 & 0.002 & 0.002 \\
Republican Party & Hope & 26 & increasing & 0.305 & 0.001 & 0.031 & 0.031 \\
\hline
\end{tabular}
 \caption{Results of the Mann–Kendall trend test for mean emotion scores by party across U.S. Congresses. Kendall's $\tau$ indicates the direction and strength of the monotonic trend, Sen's slope estimates the magnitude of change per Congress, and p-values are adjusted for multiple testing using the Benjamini–Hochberg false discovery rate procedure.
}
    \label{tab:mk_test_stats}
\end{table*}

\section{Temporal patterns in discrete emotions across CAP topics}
\begin{figure*}[hbp]
    \centering
    \includegraphics[width=\textwidth]{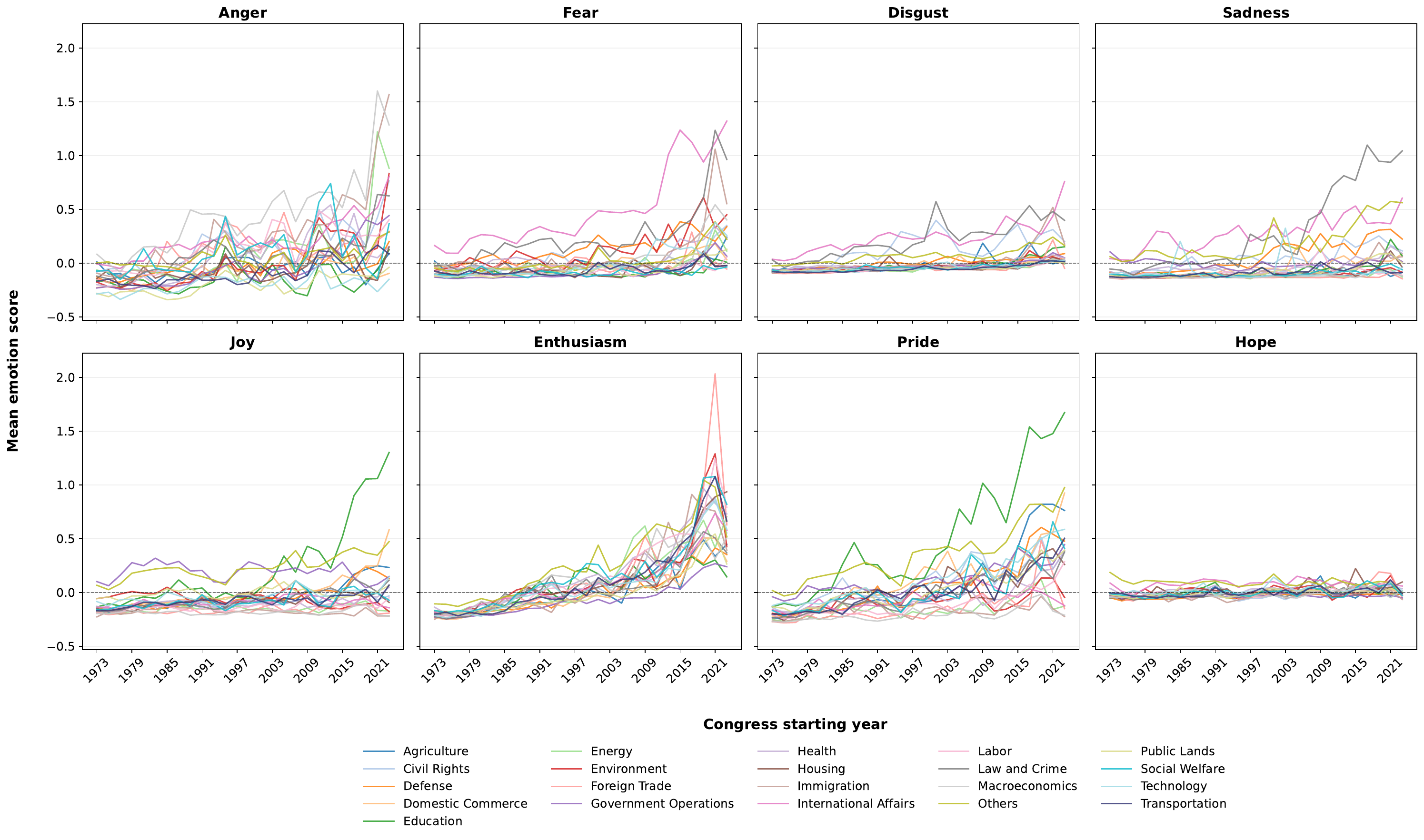}
    \caption{Trends in emotions by topic over time}
    \label{fig:emotions_trend_bytopic}
\end{figure*}

Figure \ref{fig:emotions_trend_bytopic} shows temporal patterns in each discrete emotion across CAP policy topics. Table \ref{tab:kw_test_emotion_topics} reports Kruskal--Wallis tests \cite{kruskal1952use} comparing emotion-score distributions across the 21 CAP policy topics.

\begin{table*}[htbp]
    \centering
    \begin{tabular}{lrrrrrr}
\hline
 Emotion & H & df & p & p (fdr) & $\epsilon^2$ & n \\
\hline
Anger & 60754.447 & 20 & 0.000 & 0.000 & 0.034 & 1766113 \\
Fear & 60700.367 & 20 & 0.000 & 0.000 & 0.034 & 1766113 \\
Disgust & 33088.704 & 20 & 0.000 & 0.000 & 0.019 & 1766113 \\
Sadness & 60362.656 & 20 & 0.000 & 0.000 & 0.034 & 1766113 \\
Joy & 34985.121 & 20 & 0.000 & 0.000 & 0.020 & 1766113 \\
Enthusiasm & 39777.511 & 20 & 0.000 & 0.000 & 0.022 & 1766113 \\
Pride & 39715.041 & 20 & 0.000 & 0.000 & 0.022 & 1766113 \\
Hope & 23734.278 & 20 & 0.000 & 0.000 & 0.013 & 1766113 \\
\hline
\end{tabular}
    \caption{Kruskal--Wallis tests comparing the distributions of emotion scores across the 21 CAP policy topics. P values are adjusted using the Benjamini--Hochberg false discovery rate (fdr) procedure.}
    \label{tab:kw_test_emotion_topics}
\end{table*}

\begin{table*}
\scriptsize
\centering
\begin{tabular}{lcccc}
\hline
 & Joy & Enthusiasm & Pride & Hope \\
\hline
Senate & 0.008$^{***}$ (0.007, 0.010) & -0.007$^{***}$ (-0.008, -0.006) & 0.009$^{***}$ (0.008, 0.011) & 0.001$^{***}$ (0.001, 0.001) \\
Female & 0.007$^{***}$ (0.004, 0.010) & 0.020$^{***}$ (0.017, 0.023) & 0.012$^{***}$ (0.007, 0.017) & 0.000$^{***}$ (0.000, 0.001) \\
Ideology & -0.004$^{***}$ (-0.005, -0.003) & -0.002$^{***}$ (-0.003, -0.001) & -0.006$^{***}$ (-0.008, -0.005) & -0.000$^{***}$ (-0.000, -0.000) \\
Majority party & 0.003$^{***}$ (0.003, 0.004) & 0.006$^{***}$ (0.006, 0.007) & 0.009$^{***}$ (0.009, 0.010) & 0.000$^{***}$ (0.000, 0.001) \\
Vote share & -0.000$^{**}$ (-0.001, -0.000) & -0.000$^{**}$ (-0.000, -0.000) & -0.000 (-0.001, 0.000) & 0.000 (-0.000, 0.000) \\
Republican & 0.005$^{***}$ (0.003, 0.007) & -0.002 (-0.004, 0.000) & 0.005$^{**}$ (0.002, 0.008) & -0.000$^{*}$ (-0.000, -0.000) \\ \hline
Observations & 1,766,113 & 1,766,113 & 1,766,113 & 1,766,113 \\
Marginal $R^2$ & 0.045 & 0.042 & 0.036 & 0.003 \\
Conditional $R^2$ & 0.074 & 0.129 & 0.109 & 0.005 \\

\hline
\end{tabular}
\caption{Coefficient estimates from mixed-effects models for emotions with positive valence. All models include congressional session and policy topic fixed effects. Numbers in parentheses represent 95\% confidence intervals. $^{*}p<.05$, $^{**}p<.01$, $^{***}p<.001$.}
\label{tab:positive_emotion_correlates}

\end{table*}

\begin{table*}
\centering
\scriptsize
\begin{tabular}{ccccc}
\hline
 & Anger & Fear & Disgust & Sadness \\
\hline
Senate & -0.041$^{***}$ (-0.043, -0.039) & -0.000 (-0.000, 0.000) & -0.000$^{**}$ (-0.000, -0.000) & -0.001$^{*}$ (-0.001, -0.000) \\
Female & 0.002 (-0.004, 0.008) & 0.004$^{***}$ (0.003, 0.004) & 0.000$^{***}$ (0.000, 0.000) & 0.006$^{***}$ (0.004, 0.008) \\
Ideology & 0.022$^{***}$ (0.020, 0.024) & 0.001$^{***}$ (0.000, 0.001) & 0.000$^{*}$ (0.000, 0.000) & -0.001$^{***}$ (-0.001, -0.000) \\
Majority party & -0.035$^{***}$ (-0.036, -0.034) & -0.001$^{***}$ (-0.001, -0.001) & 0.000 (-0.000, 0.000) & -0.000 (-0.000, 0.000) \\
Vote share & -0.001$^{***}$ (-0.002, -0.001) & -0.000$^{***}$ (-0.000, -0.000) & -0.000 (-0.000, 0.000) & -0.000$^{*}$ (-0.000, -0.000) \\
Republican & -0.012$^{***}$ (-0.015, -0.008) & -0.000 (-0.001, 0.000) & 0.000 (-0.000, 0.000) & -0.001 (-0.002, 0.000) \\ \hline
Observations & 1,766,113 & 1,766,113 & 1,766,113 & 1,766,113 \\
Marginal $R^2$ & 0.071 & 0.023 & 0.008 & 0.016 \\
Conditional $R^2$ & 0.124 & 0.042 & 0.014 & 0.069 \\
\hline
\end{tabular}
\caption{Mixed-effects estimates for emotions with negative valence. All models include congressional session and policy topic fixed effects. Numbers in parentheses represent 95\% confidence intervals. $^{*}p<.05$, $^{**}p<.01$, $^{***}p<.001$.}
\label{tab:negative_emotion_correlates}

\end{table*}

\begin{table*}[htbp]
\centering
\tiny
\begin{tabular}{lllll}
\hline
 & LES & LES & LES & LES \\
\hline
Anger & -0.037$^{***}$ (-0.046, -0.029) &  &  &  \\
Fear &  & -0.002 (-0.010, 0.005) &  &  \\
Disgust &  &  & -0.003 (-0.010, 0.004) &  \\
Sadness &  &  &  & -0.005 (-0.012, 0.002) \\
Ideological extremity & -0.121$^{***}$ (-0.192, -0.051) & -0.183$^{***}$ (-0.253, -0.114) & -0.184$^{***}$ (-0.253, -0.114) & -0.185$^{***}$ (-0.255, -0.116) \\
Committee chair & 0.015 (-0.014, 0.043) & 0.018 (-0.010, 0.047) & 0.018 (-0.010, 0.047) & 0.018 (-0.010, 0.046) \\
Subcommittee chair & 0.001 (-0.018, 0.019) & 0.002 (-0.017, 0.020) & 0.002 (-0.016, 0.020) & 0.002 (-0.017, 0.020) \\
Majority-party leadership & 0.101$^{***}$ (0.052, 0.149) & 0.100$^{***}$ (0.051, 0.149) & 0.100$^{***}$ (0.051, 0.149) & 0.101$^{***}$ (0.052, 0.149) \\
Minority-party leadership & -0.012 (-0.061, 0.037) & -0.006 (-0.055, 0.043) & -0.006 (-0.055, 0.043) & -0.006 (-0.055, 0.043) \\
Majority-party member & -0.033$^{***}$ (-0.051, -0.016) & -0.014 (-0.032, 0.003) & -0.014 (-0.032, 0.003) & -0.014 (-0.032, 0.003) \\
First term & -0.062$^{***}$ (-0.082, -0.043) & -0.062$^{***}$ (-0.082, -0.043) & -0.062$^{***}$ (-0.082, -0.043) & -0.063$^{***}$ (-0.082, -0.043) \\
Seniority & -0.017$^{***}$ (-0.019, -0.015) & -0.017$^{***}$ (-0.019, -0.014) & -0.017$^{***}$ (-0.019, -0.014) & -0.017$^{***}$ (-0.019, -0.014) \\
Vote share & -0.001$^{*}$ (-0.001, -0.000) & -0.001$^{*}$ (-0.001, -0.000) & -0.001$^{*}$ (-0.001, -0.000) & -0.001$^{*}$ (-0.001, -0.000) \\
Female & 0.024 (-0.009, 0.057) & 0.023 (-0.010, 0.055) & 0.022 (-0.010, 0.055) & 0.023 (-0.010, 0.056) \\
African American & -0.044 (-0.091, 0.003) & -0.034 (-0.081, 0.012) & -0.034 (-0.081, 0.012) & -0.034 (-0.081, 0.012) \\
Latino/a & 0.000 (-0.054, 0.055) & -0.001 (-0.056, 0.054) & -0.001 (-0.056, 0.053) & -0.001 (-0.055, 0.054) \\
State legislative experience & -0.002 (-0.037, 0.032) & -0.004 (-0.039, 0.030) & -0.004 (-0.038, 0.030) & -0.004 (-0.039, 0.030) \\
State leg. experience $\times$ professionalism & 0.044 (-0.055, 0.144) & 0.047 (-0.053, 0.146) & 0.046 (-0.053, 0.146) & 0.047 (-0.053, 0.147) \\
Senate & -0.220$^{***}$ (-0.251, -0.189) & -0.202$^{***}$ (-0.232, -0.171) & -0.202$^{***}$ (-0.232, -0.171) & -0.201$^{***}$ (-0.232, -0.170) \\
Republican & -0.023$^{*}$ (-0.045, -0.000) & -0.015 (-0.037, 0.008) & -0.015 (-0.037, 0.008) & -0.015 (-0.038, 0.007) \\
Number of speeches (log) & 0.072$^{***}$ (0.049, 0.094) & 0.075$^{***}$ (0.052, 0.098) & 0.075$^{***}$ (0.052, 0.098) & 0.075$^{***}$ (0.052, 0.098) \\
Speech length (log) & 0.058$^{***}$ (0.036, 0.080) & 0.052$^{***}$ (0.030, 0.074) & 0.052$^{***}$ (0.030, 0.074) & 0.052$^{***}$ (0.030, 0.073) \\ \hline
Observations & 13,671 & 13,671 & 13,671 & 13,671 \\
Marginal $R^2$ & 0.118 & 0.113 & 0.113 & 0.113 \\
Conditional $R^2$ & 0.330 & 0.326 & 0.326 & 0.326 \\
\hline
\end{tabular}
\caption{Mixed-effects model estimates predicting legislative effectiveness score (LES) using emotions with negative valence. All models include congressional session and policy topic fixed effects. Numbers in parentheses represent 95\% confidence intervals. \({}^{*}p<.05\), \({}^{**}p<.01\), and \({}^{***}p<.001\)}
\label{tab:les_negative_emotions}
\end{table*}

\begin{table*}[htbp]
\centering
\tiny

\begin{tabular}{lcccc}
\hline
 & LES & LES & LES & LES \\
\hline
Joy & 0.004 (-0.003, 0.011) &  &  &  \\
Enthusiasm &  & 0.016$^{***}$ (0.007, 0.025) &  &  \\
Pride &  &  & 0.011$^{**}$ (0.003, 0.019) &  \\
Hope &  &  &  & 0.003 (-0.003, 0.010) \\
Ideological extremity & -0.181$^{***}$ (-0.251, -0.112) & -0.171$^{***}$ (-0.241, -0.102) & -0.176$^{***}$ (-0.245, -0.106) & -0.182$^{***}$ (-0.251, -0.112) \\
Committee chair & 0.018 (-0.010, 0.047) & 0.019 (-0.009, 0.047) & 0.018 (-0.010, 0.046) & 0.019 (-0.010, 0.047) \\
Subcommittee chair & 0.002 (-0.016, 0.020) & 0.003 (-0.015, 0.022) & 0.002 (-0.017, 0.020) & 0.002 (-0.016, 0.021) \\
Majority-party leadership & 0.100$^{***}$ (0.051, 0.149) & 0.098$^{***}$ (0.049, 0.147) & 0.099$^{***}$ (0.050, 0.148) & 0.099$^{***}$ (0.050, 0.148) \\
Minority-party leadership & -0.006 (-0.055, 0.043) & -0.010 (-0.059, 0.039) & -0.007 (-0.056, 0.042) & -0.007 (-0.056, 0.043) \\
Majority-party member & -0.015 (-0.032, 0.003) & -0.019$^{*}$ (-0.036, -0.001) & -0.016 (-0.033, 0.002) & -0.015 (-0.032, 0.003) \\
First term & -0.063$^{***}$ (-0.082, -0.043) & -0.066$^{***}$ (-0.085, -0.046) & -0.063$^{***}$ (-0.082, -0.043) & -0.063$^{***}$ (-0.082, -0.043) \\
Seniority & -0.017$^{***}$ (-0.019, -0.014) & -0.016$^{***}$ (-0.019, -0.014) & -0.017$^{***}$ (-0.019, -0.014) & -0.017$^{***}$ (-0.019, -0.014) \\
Vote share & -0.001$^{*}$ (-0.001, -0.000) & -0.001$^{*}$ (-0.001, -0.000) & -0.001$^{*}$ (-0.001, -0.000) & -0.001$^{*}$ (-0.001, -0.000) \\
Female & 0.022 (-0.011, 0.054) & 0.017 (-0.015, 0.050) & 0.022 (-0.011, 0.054) & 0.022 (-0.011, 0.054) \\
African American & -0.035 (-0.081, 0.012) & -0.038 (-0.085, 0.009) & -0.036 (-0.083, 0.011) & -0.035 (-0.082, 0.012) \\
Latino/a & -0.001 (-0.056, 0.053) & -0.003 (-0.057, 0.052) & -0.003 (-0.058, 0.051) & -0.001 (-0.056, 0.053) \\
State legislative experience & -0.004 (-0.039, 0.030) & -0.002 (-0.036, 0.032) & -0.005 (-0.039, 0.029) & -0.004 (-0.038, 0.030) \\
State leg. experience $\times$ professionalism & 0.047 (-0.053, 0.147) & 0.044 (-0.056, 0.143) & 0.049 (-0.051, 0.149) & 0.046 (-0.054, 0.146) \\
Senate & -0.203$^{***}$ (-0.234, -0.172) & -0.200$^{***}$ (-0.231, -0.169) & -0.207$^{***}$ (-0.238, -0.176) & -0.202$^{***}$ (-0.233, -0.171) \\
Republican & -0.015 (-0.038, 0.007) & -0.013 (-0.035, 0.010) & -0.016 (-0.038, 0.007) & -0.015 (-0.037, 0.008) \\
Number of speeches (log) & 0.075$^{***}$ (0.052, 0.098) & 0.075$^{***}$ (0.052, 0.098) & 0.077$^{***}$ (0.054, 0.100) & 0.076$^{***}$ (0.053, 0.099) \\
Speech length (log) & 0.053$^{***}$ (0.031, 0.075) & 0.054$^{***}$ (0.032, 0.076) & 0.053$^{***}$ (0.031, 0.074) & 0.052$^{***}$ (0.030, 0.074) \\ \hline
Observations & 13,671 & 13,671 & 13,671 & 13,671 \\
Marginal $R^2$ & 0.113 & 0.114 & 0.114 & 0.113 \\
Conditional $R^2$ & 0.326 & 0.327 & 0.326 & 0.326 \\
\hline
\end{tabular}
\caption{Mixed-effects model estimates predicting legislative effectiveness score (LES) using emotions with positive valence. All models include congressional session and policy topic fixed effects. Numbers in parentheses represent 95\% confidence intervals. \({}^{*}p<.05\), \({}^{**}p<.01\), and \({}^{***}p<.001\)}
\label{tab:les_positive_emotions}
\end{table*}

\begin{table*}[htbp]
\centering
\scriptsize

\begin{tabular}{lcc}
\hline
 & LES & LES \\
\hline
Emotional valence & 0.020$^{***}$ (0.011, 0.028) &  \\
Emotional diversity & 0.026$^{***}$ (0.017, 0.034) & 0.031$^{***}$ (0.024, 0.039) \\
Emotional intensity &  & -0.014$^{**}$ (-0.025, -0.004) \\
Ideological extremity & -0.112$^{**}$ (-0.183, -0.041) & -0.130$^{***}$ (-0.201, -0.060) \\
Committee chair & 0.015 (-0.013, 0.043) & 0.015 (-0.013, 0.044) \\
Subcommittee chair & 0.002 (-0.017, 0.020) & 0.001 (-0.017, 0.020) \\
Majority-party leadership & 0.096$^{***}$ (0.047, 0.145) & 0.099$^{***}$ (0.050, 0.147) \\
Minority-party leadership & -0.017 (-0.066, 0.032) & -0.014 (-0.063, 0.034) \\
Majority-party member & -0.033$^{***}$ (-0.051, -0.016) & -0.029$^{**}$ (-0.046, -0.011) \\
First term & -0.068$^{***}$ (-0.087, -0.048) & -0.066$^{***}$ (-0.086, -0.047) \\
Seniority & -0.017$^{***}$ (-0.019, -0.014) & -0.017$^{***}$ (-0.019, -0.014) \\
Vote share & -0.001$^{*}$ (-0.001, -0.000) & -0.001$^{*}$ (-0.001, -0.000) \\
Senate & -0.217$^{***}$ (-0.248, -0.186) & -0.208$^{***}$ (-0.238, -0.177) \\
Republican & -0.022 (-0.044, 0.001) & -0.021 (-0.043, 0.002) \\
Female & 0.016 (-0.017, 0.048) & 0.018 (-0.015, 0.051) \\
African American & -0.047$^{*}$ (-0.094, -0.000) & -0.044 (-0.091, 0.003) \\
Latino/a & -0.005 (-0.059, 0.050) & -0.001 (-0.056, 0.053) \\
State legislative experience & -0.002 (-0.037, 0.032) & -0.002 (-0.036, 0.032) \\
State experience $\times$ professionalism & 0.047 (-0.052, 0.147) & 0.043 (-0.056, 0.143) \\
Number of speeches (log) & 0.076$^{***}$ (0.053, 0.099) & 0.078$^{***}$ (0.055, 0.100) \\
Speech length (log) & 0.051$^{***}$ (0.029, 0.073) & 0.044$^{***}$ (0.022, 0.066) \\ \hline
Observations & 13,671 & 13,671 \\
Marginal $R^2$ & 0.119 & 0.118 \\
Conditional $R^2$ & 0.331 & 0.330 \\
\hline
\end{tabular}
\caption{Mixed-effects model estimates predicting legislative effectiveness score (LES) using broader characteristics of emotional expression. All models include congressional session and policy topic fixed effects. Numbers in parentheses represent 95\% confidence intervals. \({}^{*}p<.05\), \({}^{**}p<.01\), and \({}^{***}p<.001\)}
\label{tab:les_broader_emotions}
\end{table*}

\end{document}